\documentclass[aoas,MSNbibl,nameyear,seceqn,dvips]{arximspdf}
\usepackage{algorithm}
\usepackage{algorithmic}
\usepackage{graphicx}

\doi{10.1214/12-AOAS549} 
\volume{6}
\issue{3}
\pubyear{2012}
\firstpage{1095}
\lastpage{1117}

\makeatletter

\newproclaim{ex}{Example}

\newcommand{\argmin}{\mathop{\arg\min}}
\newcommand{\bX}{\mathbf{X}}
\newcommand{\bY}{\mathbf{Y}}
\newcommand{\bB}{\mathbf{B}}

\newcommand{\Tr}{\operatorname{Tr}}
\newcommand{\blbbeta}{\bolds{\beta}}

\newcommand{\bu}{{\mathbf u}}
\newcommand{\bW}{{\mathbf W}}
\newcommand{\bA}{{\mathbf A}}
\newcommand{\bV}{{\mathbf V}}

\newcommand{\Vint}{V_{\mathrm{int}}}
\newcommand{\Vleaf}{V_{\mathrm{leaf}}}

\makeatother

\begin{document}
\begin{frontmatter}

\title{Tree-guided group lasso for multi-response regression with structured
sparsity, with~an~application to eQTL mapping\thanksref{T1}}
\runtitle{Tree lasso for eQTL mapping}

\thankstext{T1}{Supported in part by NIH 1R01GM087694.}

\begin{aug}
\author[A]{\fnms{Seyoung} \snm{Kim}\ead[label=e1]{sssykim@cs.cmu.edu}}
\and
\author[A]{\fnms{Eric P.} \snm{Xing}\corref{}\thanksref{t2}\ead[label=e2]{epxing@cs.cmu.edu}}
\runauthor{S. Kim and E. P. Xing}
\affiliation{Carnegie Mellon University}
\address[A]{School of Computer Science \\
Carnegie Mellon University \\
Pittsburgh, Pennsylvania 15213 \\
USA \\
\printead{e1}\\
\hphantom{E-mail: }\printead*{e2}} 
\end{aug}

\thankstext{t2}{Supported in part by ONR N000140910758,
NSF DBI-0640543, NSF CCF-0523757
and an Alfred P. Sloan Research Fellowship.}

\received{\smonth{12} \syear{2009}}
\revised{\smonth{2} \syear{2012}}

%
\begin{abstract}
We consider the problem of estimating a sparse multi-response
regression function,
with an application to expression quantitative trait locus (eQTL) mapping,
where the goal is to discover
genetic variations that influence gene-expression levels.
In particular, we investigate a shrinkage technique capable of
capturing a given hierarchical
structure over the responses, such as a hierarchical clustering tree
with leaf nodes for responses and internal nodes for clusters of
related responses at multiple granularity, and we seek to leverage this
structure to recover covariates relevant to each hierarchically-defined
cluster of responses.
We propose a tree-guided group lasso, or \textit{tree lasso},
for estimating such structured sparsity under multi-response regression
by employing a novel penalty function constructed from the tree.
We describe a systematic weighting scheme for the overlapping
groups in the tree-penalty such that each regression coefficient is penalized
in a balanced manner despite the inhomogeneous multiplicity of group
memberships of the regression coefficients due to overlaps among groups.
For efficient optimization, we employ a smoothing proximal gradient method
that was originally developed for a general class of
structured-sparsity-inducing penalties.
Using simulated and yeast data sets, we demonstrate that our
method shows a superior performance in terms of both prediction errors
and recovery of true sparsity patterns, compared to other methods
for learning a~multivariate-response regression.
%
%
%
\end{abstract}

%

\end{frontmatter}

\section{Introduction}\label{sec1}

Recent advances in high-throughput technology for profiling gene
expressions and assaying genetic variations at a genome-wide scale have provided
researchers an unprecedented opportunity to comprehensively study
the
genetic causes of complex diseases
such as asthma, diabetes, and cancer.
\textit{Expression quantitative trait locus} (eQTL) mapping
considers gene expression measurements, also known as
\textit{gene-expression traits}, as intermediate phenotypes, and aims to
identify the genetic markers such as single nucleotide polymorphisms
(SNPs) that influence the expression levels of genes,
which gives rise to the variability in clinical phenotypes or disease
susceptibility across individuals. This type of analysis can provide a
deeper insight
into the functional role of the eQTLs
in the disease process by linking the SNPs to genes whose functions are
often known directly or
indirectly through other co-expressed genes in the same pathway.

The most commonly used method for eQTL analysis has been to examine the
expression level of a single gene at a time
for association, treating genes as independent of each other
[\citet{Cheung2005}, \citet{Str2005}, \citet{brem2008}].
However, it is widely believed that many of the genes in the same
biological pathway
are often co-expressed or co-regulated [\citet{pujana2008}, \citet{Zhang2005}]
and may share a common genetic basis that causes the variations in
their expression levels.
How to incorporate such information on relatedness of genes into
statistical analysis
of associations between SNPs and gene expressions remains an
under-addressed problem.
One of the popular existing approaches is to consider the relatedness
of genes \textit{after}
rather than \textit{during} statistical analysis of eQTL data,
which obviously fails to fully exploit the statistical power from this
additional source of information.
Specifically, in order to find the genetic variations with pleiotropic
effects that perturb
the expressions of multiple related genes jointly,
in recent eQTL studies, the expression traits for individual genes were
analyzed separately,
and then the results were examined for all genes
in light of gene modules to see if any gene sets are enriched for
association with a common SNP [\citet{brem2008}, \citet{schadtdecode},
\citet{schadtmouse}].
This type of analysis uses the information on gene
modules only in the post-processing step after a set of single-gene analyses,
instead of directly incorporating the correlation pattern in gene
expressions in the process
of searching for SNPs with pleitropic effects.

Recently, a different approach for searching for SNPs with pleiotropic
effects has been proposed
to leverage information on gene modules more directly
[\citet{Segal2003}, \citet{koller2006}].
In this approach, the module network originally developed for
discovering clusters of co-regulated
genes from gene expression data was extended to include SNPs as
potential regulators that can influence
the activity of gene modules.
The main weakness of this method is that it computed the averages of
gene-expression levels
over those genes within each module and looked for SNPs that affect the
average gene expressions of the module.
The operation of computing averages
can lead to a significant loss of information on the detailed
activity of individual genes and negative correlations within a module.

%
\begin{figure}
\begin{tabular}{@{}c@{\qquad}c@{}}

\includegraphics{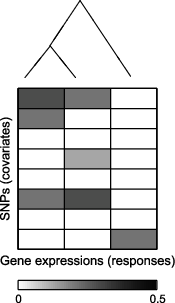}
 & \includegraphics{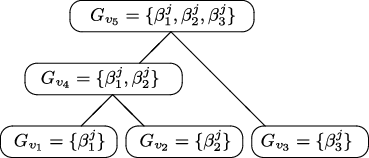}\\
(a) & (b)
\end{tabular}
\caption{An illustration of a tree lasso. \textup{(a)}: The sparse
structure in regression coefficients is shown with white entries for
zeros and gray entries for nonzero values. The hierarchical clustering
tree represents the correlation structure in responses. The first two
responses are highly correlated according to the clustering tree, and
are likely to be influenced by the same covariates. \textup{(b)}: Groups of
variables associated with each node of the tree in panel \textup{(a)} in the
tree-lasso penalty.} \label{figillustt}\vspace*{1pt}
\end{figure}

In this article we propose a tree-guided group lasso, or tree lasso,
that directly combines statistical strength across multiple related
genes in gene expression data to identify SNPs with pleiotropic effects
by leveraging any given knowledge of a hierarchical clustering tree
over genes.\setcounter{footnote}{2}\footnote{Here we focus on making
use of the given knowledge of related genes to enhance the power of
eQTL analysis, rather than discovering or evaluating how genes are
related, which are interesting problems in their own right, and are
studied widely [\citet{Segal2003}]. If the gene co-expression
pattern is not available, one can simply run any off-the-shelf
hierarchical agglomerative clustering algorithm on the gene-expression
data to obtain one before applying our method. It is beyond the scope
of this paper to discuss, compare, and further develop such algorithms
for clustering genes or learning trees.\looseness=1} The hierarchical clustering
tree contains clusters of genes at multiple granularity, and genes
within a cluster have correlated expression levels. The leaf nodes of
the tree correspond to individual genes, and each internal node
represents a cluster of genes at the leaf nodes of the subtree rooted
at the internal node in question. Furthermore, each internal node in
the tree is associated with a weight that represents the height of the
subtree, or how tightly the genes in the cluster for that internal node
are correlated. As illustrated in Figure~\ref{figillustt}(a), the
expression levels of genes in each cluster are likely to be influenced
by a common set of SNPs, and this type of sharing of genetic effects
among correlated genes is stronger among tightly correlated genes in
the cluster at the lower levels with a smaller height in the tree than
among loosely correlated genes in the cluster near the root of the tree
with a greater height. This multi-level grouping structure of genes can
be available either as prior knowledge from domain experts, or can be
learned from the gene-expression data using various clustering
algorithms such as the hierarchical agglomerative clustering algorithm
[\citet{Golub1999}].

Our method is based on a multivariate regression method with a
regularization function that is constructed from the hierarchical
clustering tree. This regularizer induces a structured shrinkage effect
that encourages multiple correlated responses to share a similar set of
relevant covariates, rather than having independent sets of relevant
covariates. This is a biologically and statistically desirable bias not
present in existing methods for identifying eQTLs. For example,
assuming that the SNPs are represented as covariates, gene expressions
as responses, and the association strengths as regression coefficients
in a regression model, a multivariate regression with an $L_1$
regularization, called the lasso, has been applied to identify a small
number of SNPs with nonzero association strengths [\citet{TongTong2009}].
Here, the lasso treats multiple responses as independent of each other
and selects relevant covariates for each response variable separately.
Although the $L_1$ penalty in the lasso can be extended to the
$L_1/L_2$ penalty, also known as the group-lasso penalty, for union
support recovery, where all of the responses are constrained to have
the same relevant covariates [\citet{Obo2008a}, \citet{Obo2008b}], in this
case, the rich and heterogeneous relatedness among the responses as
captured by a~weighted tree cannot be taken into account.

Our method extends the $L_1/L_2$ penalty
to the tree-lasso penalty by letting the hierarchically-defined groups
overlap. The
tree-lasso penalty achieves \textit{structured sparsity},
where the related responses (i.e., gene expressions) in the same group share
a common set of relevant covariates (i.e., SNPs), in a~way that is
properly calibrated to
the strength of their relatedness and consistent with their overlapping
group organization.
Although several schemes have been previously proposed to use the
group-lasso penalty with overlapping
groups to take advantage of a more complex structural information on
response variables,
due to their \textit{ad hoc} weighting scheme for different overlapping
groups in the regularization function,
some regression coefficients were penalized arbitrarily more heavily
than others, leading
to an inconsistent estimate [\citet{Zhao2009}, \citet{Jac2009},
\citet{Bach2009}].
In contrast, we propose a systematic weighting scheme for overlapping groups
that applies a balanced penalization to all of the regression coefficients.
Since the tree lasso is a special case of overlapping group lasso, where
the weights and overlaps of groups are determined according to the
hierarchical clustering tree,
we adopt for efficient optimization
the smoothing proximal gradient (SPG) method [\citet{Chen2011}]
that was developed for optimizing a convex loss function with a general
class of
structured-sparsity-inducing penalty functions including
overlapping group lasso.

Compared to our previous work on the graph-guided fused lasso that
leverages a network structure over responses to achieve structured sparsity
[\citet{Kim2009b}], the tree lasso has a considerably
lower computational time, and allows more than thousands of response variables
to be analyzed simultaneously as is necessary in a typical eQTL mapping.
This is in part because the computation time in the graph-guided fused
lasso depends
on the number of edges in the graph
that can be as large as $|V|\times|V|$, where $|V|$ is the number of
response variables, whereas
in the tree lasso, it is determined by the number of nodes in the tree,
which is bounded by twice the number of response variables.
Another potential advantage of the tree lasso
is that it relaxes the constraint in the graph-guided fusion penalty that
the regression coefficients should take the similar values for a
covariate relavant to multiple correlated responses.
Although introducing this bias through
the fusion penalty in the graph-guided fused lasso offered the benefit of
combining weak association signals and reducing false positives, it is expected
that relaxing this constraint
could further increase the power. The $L_1/L_2$ penalty in our tree
regularization
achieves a joint selection of covariates for multiple related responses,
while allowing different values for
the regression coefficients corresponding to the selected covariate and
correlated response variables.

Although the hierarchical agglomerative clustering algorithm has been
widely popular as a preprocessing step for regression or classification
tasks [\citet{Golub1999}, \citet{Sor2001}, \citet{Hastie2001}], our proposed method is
the first to make use of the full results from the clustering algorithm
given as tree structure and subtree-height information. Most of the
previous classification or regression methods that build on the
hierarchical clustering algorithm used summary statistics extracted
from the hierarchical clustering tree such as subsets of genes forming
clusters or averages of gene expressions within each cluster, rather
than using the tree as it is [\citet{Golub1999}, \citet{Hastie2001}]. In the
tree lasso, we use the full hierarchical clustering tree as prior
knowledge to construct a regularization function. Thus, the tree lasso
incorporates the full information present in both the raw data and the
hierarchical clustering tree to maximize the power for detecting weak
association signals and to reduce false positives. In our experiments,
we demonstrate that our proposed method can be successfully applied to
select SNPs affecting the expression levels of multiple genes, using
both simulated and yeast data sets.

The remainder of the paper is organized as follows. In Section~\ref{sec2} we provide
a brief discussion of previous work on sparse regression estimation. In
Section~\ref{sec3}
we introduce the tree lasso and describe an efficient optimization
method based on
SPG. We present experimental results on simulated and yeast eQTL data
sets in Section~\ref{sec4},
and conclude in Section~\ref{sec5}.

\section{Background on multivariate regression approach for eQTL mapping}\label{sec2}


Let us assume that data are collected for $J$ SNPs and $K$
gene-expression traits over $N$ individuals.
Let $\mathbf{X}$ denote the $N\times J$ matrix of SNP genotypes for
covariates, and
$\mathbf{Y}$ the $N\times K$ matrix of gene-expression measurements
for responses.
In eQTL mapping,
each element of the $\mathbf{X}$ takes values from $\{0,1,2\}$
according to the number of minor alleles at the given locus in each individual.
Then, we assume a linear model for the functional mapping from
covariates to
response variables:
%
\begin{equation}\label{eqm}
\mathbf{Y} = \mathbf{X}\mathbf{B}+\mathbf{E},
\end{equation}
where $\mathbf{B}$ is the $J\times K$ matrix of regression coefficients
and $\mathbf{E}$ is the $N\times K$ matrix of noise terms distributed
as mean 0 and a constant variance. We center each column of $\mathbf
{X}$ and $\mathbf{Y}$
such that the mean is zero, and consider the model without an intercept.
Throughout this paper, we use subscripts and superscripts to denote
rows and columns of a matrix, respectively
(e.g., $\blbbeta_j$ and $\blbbeta^k$ for
the $j$th row and $k$th column of $\mathbf{B}$).


When $J$ is large and the number of relevant covariates is small,
the lasso offers an effective method for identifying the small number of
nonzero elements in $\bB$ [\citet{lasso}].
The lasso obtains $\hat{\mathbf{B}}^{\mathrm{lasso}}$ by solving the
following optimization problem:
%
\begin{equation} \label{eqlasso}
\hat{\mathbf{B}}^{\mathrm{lasso}} = \argmin_\bB
\frac{1}{2}{\| \mathbf{Y} - \mathbf{X}\mathbf{B} \|}_F^2
+ \lambda\|\bB\|_1,
\end{equation}
where $\| \cdot\|_F$ is the Frobenius norm,
$\| \cdot\|_1$ is the matrix $L_1$ norm, and $\lambda$ is
a tuning parameter that controls the amount of sparsity
in the solution. Setting $\lambda$ to a small value leads to a smaller
number of nonzero regression coefficients.


The lasso estimation in (\ref{eqlasso}) is equivalent to
selecting relevant covariates for each of the $K$ responses separately,
and does not provide any mechanism to enforce a joint selection of
common relevant covariates for multiple related responses. In the
literature of multi-task learning, an $L_1/L_2$ penalty, also known as
a group lasso penalty [\citet{glasso}], has been adopted in
multivariate-response regression to take advantage of the relatedness
of the response variables and recover the union support---the pattern
of nonzero regression coefficients shared across all of the
responses [\citet{Obo2008a}]. This method is widely known as the
$L_1/L_2$-regularized multi-task regression in the machine learning
community, and its estimate for regression coefficients is given as
%
\begin{equation}\label{eqL1L2}
\hat{\mathbf{B}}^{L_1/L_2} = \argmin_\bB
\frac{1}{2}{\| \mathbf{Y} - \mathbf{X}\mathbf{B} \|}_F^2
+ \lambda\sum_j {\Vert\blbbeta_{j} \Vert}_2,
\end{equation}
where $\|\cdot\|_2$ denotes an $L_2$ norm.
In $L_1/L_2$-regularized multi-task regression,
an $L_2$ norm is applied to the regression coefficients for all
responses for each covariate,\vadjust{\goodbreak} $\blbbeta_j$, and these $L_2$ norms
for the $J$ covariates
are combined through an $L_1$ norm to encourage only a small number of
covariates to take
nonzero regression coefficients.
Since the $L_2$ part of the penalty does not have the property of
encouraging sparsity,
if the $j$th covariate is selected as relevant, then all of the
elements of $\blbbeta_j$ would take nonzero values, although the
regression coefficient values for the covariate
are still allowed to vary across different responses.
When applied to eQTL mapping, this method is significantly limited
since it is not realistic to assume that the expression levels of
all of the genes are influenced by the same set of relevant SNPs.
A subset of co-expressed
genes may be perturbed by a common set of SNPs, and genes in a
different pathway
are less likely to be affected by the same SNPs.
The sparse group lasso [\citet{Friedman2010}] can be adopted to relax
this constraint
by adding a lasso penalty to (\ref{eqL1L2}) so that
individual regression
coefficients within each $L_2$ norm can be set to zeros. However, this method
shares the same limitation as the $L_1/L_2$-regularized multi-task
regression in
that it cannot incorporate complex
grouping structures in the responses such as groups at multiple
granularity as in the hierarchical clustering tree.


\section{Tree lasso for exploiting hierarchical clustering tree in
eQTL mapping}\label{sec3}

We introduce the tree lasso that considerably adds
flexibility and power to these existing methods by taking advantage of
the complex correlation
structure given as a hierarchical clustering tree over the responses.
We present a~highly efficient algorithm for estimating the parameters
in a tree lasso that is based on the smoothing proximal gradient
descent developed for a general class of structured-sparsity-inducing norms.

\subsection{Tree lasso}\label{sec31}

In a microarray experiment, gene-expression levels are measured for
more than thousands of genes at a time, and many of the genes show
highly correlated expression levels across samples, implying they may
share a common regulator or participate in the same pathway. In
addition, in eQTL analysis, it is widely believed that genetic
variations such as SNPs perturb modules of related genes rather than
acting on individual genes. As these gene modules are often derived and
visualized by running the hierarchical agglomerative clustering
algorithm on gene expression data, a natural extension of sparse
regression methods for eQTL mapping is to incorporate with them the
output of the hierarchical clustering algorithm to identify genetic
variations that influence gene modules in the clustering tree. In this
section, we build on the $L_1/L_2$-regularized regression and introduce
a~tree lasso that can directly leverage hierarchically-organized groups
of genes to combine statistical strength across the expression levels
of genes within each group. Although our work is primarily motivated by
eQTL mapping in genetics, the tree lasso is generally applicable to any
multivariate-response regression problem, where the hierarchical group
structure over the responses is given as desirable sources of\vadjust{\goodbreak}
structural bias, such as in many computer vision
[\citet{Yuan2010}] and natural language processing applications
[\citet{Zhang2010}, \citet{Zhou2010}], where dependencies
among visual objects and among parts of speech are well known to be
valuable to enhance prediction performance.

Assume that the relationship among the $K$ responses is represented as
tree $T$
with a set of vertices $V$ of size $|V|$.
As illustrated in Figure~\ref{figillustt}(a), each of the $K$ leaf nodes is
associated with a response variable, and each of the internal nodes represents
a group of the responses located at the leaves of the subtree rooted at the
given internal node.
Internal nodes near
the bottom of the tree correspond to tight clusters of highly related responses,
whereas the internal nodes near the root represent
groups with weak correlations among the responses in its subtree.
This tree structure may be provided as prior knowledge by domain experts
or external resources (e.g., gene ontology databases in our eQTL
mapping problem),
or can be learned from the data for response variables using methods
such as
the hierarchical agglomerative clustering algorithm. We assume that
each node $v \in V$ of the tree is associated with height $h_v$
of the subtree rooted at $v$, representing how tightly its members are
correlated.
In addition, we assume that the heights~$h_v$'s of the internal nodes are
normalized so that the height of the root node is 1.

Given this tree $T$ over the $K$ responses, we generalize the $L_1/L_2$
regularization
in (\ref{eqL1L2}) to a tree regularization
by expanding the $L_2$ part of the $L_1/L_2$ penalty into an
overlapping group lasso penalty.
The overlapping groups in tree regularization are defined based on tree
$T$ as follows.
Each node $v\in V$ of tree $T$
is associated with group $G_v$ whose members are the response variables
at the leaf nodes of the subtree rooted at node $v$. For example,
Figure~\ref{figillustt}(b) shows the groups of responses and the corresponding
regression coefficients that are associated with each of the nodes of the
tree in Figure~\ref{figillustt}(a). Given these overlapping groups,
we define the tree lasso as
%
\begin{equation}\label{eqT}
\hat{\mathbf{B}}^{T} = \argmin_\bB
\frac{1}{2}{\| \mathbf{Y} - \mathbf{X}\mathbf{B} \|}_F^2
+ \lambda\sum_j \sum_{v \in V} w_v {\Vert\blbbeta_j^{G_v} \Vert
}_2,
\end{equation}
where $\blbbeta_j^{G_v}$ is a vector of regression coefficients
$\{\beta_j^k | k\in G_v \}$. Since a tree associated with $K$ responses
can have at most $2K$ nodes, the number of $L_2$ terms that appear in
the tree-lasso
penalty is upper-bounded by $|V| = 2K$ for each covariate.



Each group\vspace*{1pt} of regression coefficients $\blbbeta_j^{G_v}$ in
(\ref{eqT})
is weighted with $w_v$ such that the group of responses near the leaf
of the tree is more
likely to have common relevant covariates,
while ensuring the amount of penalization aggregated over all of the
overlapping groups
for each regression coefficient to be the same for all regression coefficients.
We define $w_v$'s in (\ref{eqT}) in terms of two quantities
$g_v$'s and $s_v$'s,
given as $s_v=h_v$ and $g_v=1-h_v$, that are associated with each
internal node $v$ of
height $h_v$ in tree $T$.
The~$s_v$ represents the weight for selecting relevant covariates
separately for
the responses associated with each child of node $v$,
whereas the $g_v$
represents the weight for selecting relevant covariates jointly for the
responses
for all of the children of node $v$.
We first consider a simple
case with two responses ($K=2$) and a~tree of three nodes that consists
of two leaf nodes
($v_1$ and $v_2$) and one root node ($v_3$), and then generalize this
to an arbitrary
tree. When $K=2$, the penalty term in (\ref{eqT}) can be
written as
%
\begin{equation} \label{eqenet}
\sum_j \sum_{v \in V} w_v {\Vert\blbbeta^j_{G_v}
\Vert}_2
= \sum_j \bigl[ s_3 (|\beta^j_1|+|\beta^j_2| )
+ g_3 \bigl(\sqrt{(\beta^j_1)^2+(\beta^j_2)^2}\bigr) \bigr],
\end{equation}
where the group weights are set to $w_{v_1}=s_3$, $w_{v_2}=s_3$, and
$w_{v_3}=g_3$. Equation (\ref{eqenet}) has a similar form to the
elastic-net penalty [\citet{Zou2005}], with the slight difference that
the elastic net uses the square of the~$L_2$~norm. The $L_1$ norm and
$L_2$ norm in (\ref{eqenet}) are weighted by $s_3$ and~$g_3$,
and play the role of setting $\beta_j^1$ and $\beta_j^2$ to nonzero
values separately or jointly. A large value of $g_v$ indicates that the
responses are highly related, and a~joint covariate selection is
encouraged by heavily weighting the $L_2$ part of the penalty. When
$s_3=0$, the penalty in (\ref{eqenet}) is equivalent to the
$L_1/L_2$-regularized multi-task regression in
(\ref{eqL1L2}), where the responses share the same set of relevant
covariates, whereas setting $g_3=0$ in (\ref{eqenet}) leads
to a~lasso penalty. In general, given a~single-level tree with all of
the responses under a single parent node, the tree-lasso penalty
corresponds to a linear combination of $L_1$ and $L_2$ penalties as in
(\ref{eqenet}).

\begin{figure}
\begin{tabular}{@{}c@{}c@{}c@{}}

\includegraphics{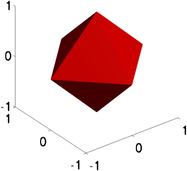}
 & \includegraphics{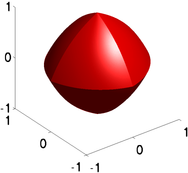} & \includegraphics{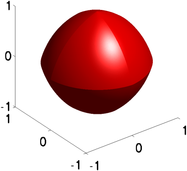}\\
(a) & (b) & (c)\\[4.5pt]
\end{tabular}
\begin{tabular}{@{}cc@{}}

\includegraphics{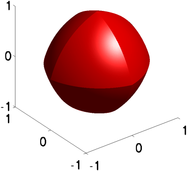}
 & \includegraphics{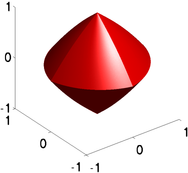}\\
(d) & (e)
\end{tabular}
\caption{Unit contour\vspace*{1pt} surfaces for $\{\beta^1_j, \beta^2_j,
\beta^3_j\}$ in various penalties, assuming the tree structure over
responses in Figure \protect\ref{figillustt}. \textup{(a)}: Lasso,
\textup{(b)}: tree lasso with $g_1=0.5$ and $g_2=0.5$, \textup{(c)}:
$g_1=0.7$ and $g_2=0.7$, \textup{(d)}: $g_1=0.2$ and $g_2=0.7$, and
\textup{(e)}: $g_1=0.7$ and $g_2=0.2$.}\vspace*{0.5pt} \label{figconst}
\end{figure}

Now, we generalize this process of obtaining $w_v$'s in the tree-lasso
penalty for the
special case of a single-level tree to an arbitrary tree.
Starting from the root node and traversing down the tree recursively to
the leaf nodes,
at each of the root and internal nodes,
we apply the similar operation of linear combination of the $L_1$ norm
and $L_2$ norm as in (\ref{eqenet}) as follows:
%
\begin{equation}\label{eqTsg}
\sum_j \sum_{v \in V} w_v {\Vert\blbbeta_j^{G_v} \Vert}_2
= \sum_j W_j(v_{\mathrm{root}}),
\end{equation}
where
\[
W_j(v) = \cases{
\displaystyle s_v \cdot{\sum_{c \in\mathrm{Children}(v)}} |W_j(c)| +
g_v \cdot{\Vert\blbbeta_j^{G_v} \Vert}_2,
&\quad if $v$ is an internal node,\vspace*{2pt}\cr
\displaystyle {\sum_{m \in G_v}}|\beta_j^{m}|,
&\quad if $v$ is a leaf node.}
%
\]
Then, it can be shown that the following relationship holds between
$w_v$'s and ($s_v$, $g_v$)'s:
\[
w_v = \cases{
\displaystyle g_v {\prod_{m \in\mathrm{Ancestors}(v)}} s_m,
&\quad if $v$ is an internal node,\vspace*{2pt}\cr
\displaystyle {\prod_{m \in\mathrm{Ancestors}(v)}} s_m,
&\quad if $v$ is a leaf node.}
\]
The above weighting scheme extends the linear combination of the $L_1$
and~$L_2$ penalty
in (\ref{eqenet}) hierarchically,
so that the $L_1$ and~$L_2$ norms encourage separate and joint
selections of
covariates for the given groups of responses. The $s_v$'s and $g_v$'s
determine the balance between these $L_1$ and $L_2$ norms.
If $s_v=1$ and $g_v=0$ for all $v \in V$, then only separate selections
are performed,
and the tree-lasso penalty reduces to the lasso penalty.
On the other hand, if $s_v=0$ and $g_v=1$ for all $v \in V$, the penalty
reduces to the $L_1/L_2$ penalty in (\ref{eqL1L2}) that
constrains all of the responses to have the same set of relevant covariates.
The unit contour surfaces of various penalties for $\beta_j^1$, $\beta
_j^2$, and $\beta_j^3$
with groups as defined in Figure~\ref{figillustt}
are shown in Figure~\ref{figconst}.

%

The seemingly complex method for determining the weights $w_v$'s for
groups in the tree-lasso penalty has the property of ensuring all of
the regression coefficients to be overall penalized by an equal amount
across all nested overlapping groups as they appear in a balanced
manner. Proposition~1 (as stated and proved in the supplemental article
[\citet{Kim12}]) shows that even if each response $k$ belongs to
multiple groups associated with different internal nodes $\{v\dvtx k
\in G_v\}$ and appears multiple times in the overall penalty in
(\ref{eqTsg}), the sum of weights over all of the groups that contain
the given response is always one. Thus, the weighting scheme in
(\ref{eqTsg}) guarantees that all of the individual regression
coefficients are overall penalized equally. Although several variations
of group lasso with overlapping groups have been proposed previously,
all of those methods weighted the $L_2$ norms for overlapping groups
with arbitrarily defined weights, resulting in unbalanced weights for
different regression coefficients [\citet{Zhao2009},
\citet{Bach2009}]. It was empirically shown that these arbitrary
weighting schemes give an inconsistent estimate [\citet{Bach2009}].


Below, we provide an example of
the process of constructing a tree-lasso penalty based on the simple tree
over three responses in Figure~\ref{figillustt}(a). For more
complex trees over a large number of responses, the same procedure
can be applied, traversing the tree recursively from the root to the
leaf nodes.

\begin{ex} Given the tree in Figure~\ref{figillustt}, for the $j$th
covariate the penalty of
the tree lasso in (\ref{eqTsg}) can be written as follows:
\begin{eqnarray*}
W_j(v_1) &=& |\beta_j^{1}|,\qquad
W_j(v_2) = |\beta_j^{2}|,\qquad W_j(v_3) = |\beta_j^{3}|,
\\[-2pt]
W_j(v_4) &=& g_{v_4}\cdot{\Vert\blbbeta_j^{G_{v_4}}
\Vert}_2
+s_{v_4}\cdot\bigl(|W_j(v_1)|+|W_j(v_2)|\bigr)
\\[-2pt]
&=& g_{v_4}\cdot{\Vert\blbbeta_j^{G_{v_4}} \Vert}_2
+s_{v_4}\cdot(|\beta_j^{1}|+|\beta_j^{2}|),
\\[-2pt]
W_j(v_{\mathrm{root}}) &=& W_j(v_5)
= g_{v_5}\cdot{\Vert\blbbeta_j^{G_{v_5}} \Vert}_2
+s_{v_5}\cdot\bigl(|W_j(v_4)|+|W_j(v_3)|\bigr)
\\[-2pt]
&=& g_{v_5}\cdot{\Vert\blbbeta_j^{G_{v_5}} \Vert}_2
+s_{v_5}\cdot g_{v_4} {\Vert\blbbeta_j^{G_{v_4}} \Vert}_2
+ s_{v_5}\cdot s_{v_4}(|\beta_j^{1}|+|\beta_j^{2}|)+s_{v_5}|\beta_j^{3}|.
\end{eqnarray*}
\end{ex}

The tree-lasso penalty that we introduced above can be easily extended
to other related types of structures such as trees with different
branching factors and a forest that consists of multiple trees.
In addition, our proposed regularization can be applied to
a pruned tree whose leaf nodes contain groups of variables instead of individual
variables.

\subsection{Parameter estimation}\label{sec32}

Although the tree-lasso optimization problem in (\ref{eqT})
is convex, the main challenges for solving equation (\ref{eqT}) arise
from the nonseparable $L_2$ terms over ${\bolds\beta}_g^{G_v}$'s
in the nonsmooth
penalty. While the coordinate descent algorithm has been successfully
applied to nonsmooth penalties such as the lasso and group lasso with
nonoverlapping groups [\citet{Friedman07}], it cannot be applied to the
tree lasso because the overlapping groups with nonseparable terms in
the penalty prevent us from obtaining a closed-form update equation for
iterative optimization. While the optimization problem for the tree
lasso can be formulated as a second-order cone program and solved with
the interior point method [\citet{Boyd2004}], this approach does not
scale to high-dimensional problems such as eQTL mapping that involves a
large number of SNPs and gene-expression measurements. Recently, a
smoothing proximal gradient (SPG) method was developed for an efficient
optimization of a convex loss function with a general class of
structured-sparsity-inducing penalty functions that share the same
challenges of nonsmoothness and nonseparability [\citet{Chen2011}]. The\vadjust{\goodbreak}
SPG can handle a wide variety of penalties such as the overlapping
group lasso and fused lasso, and as the tree lasso is a~special case of
the overlapping group lasso, we adopt this method in our paper. As we
detail below in this section, SPG first decouples the nonseparable
terms in the penalty by reformulating it with a~dual norm, and
introduces a~smooth approximation of the nonsmooth penalty.
Then, in order to optimize the objective function with this smooth
approximation of the penalty, SPG adopts the fast iterative shrinkage
thresholding algorithm (FISTA) [\citet{FISTA}], an accelerated
gradient descent method, to optimize the objective function
an accelerated gradient descent method.\looseness=-1

\subsubsection{Reformulation of the penalty function}

We rewrite (\ref{eqT}) by splitting the tree-lasso penalty into
two parts corresponding to two sets of nodes in tree $T$,
$\Vint=\{v| |G_v| >1 \}$ for all of the internal nodes
and $\Vleaf=\break\{v| |G_v| = 1 \}$ for all of the leaf nodes, as follows:
%
\begin{eqnarray}
\label{eqpsplit}
\hat{\mathbf{B}}^{T} &=& \argmin
\frac{1}{2}{\| \mathbf{Y} - \mathbf{X}\mathbf{B} \|}_F^2
+ \lambda\sum_{j=1}^J \sum_{v \in\Vint} w_v {\Vert
\blbbeta_j^{G_v} \Vert}_2 \nonumber\\[-9pt]\\[-9pt]
&&{}+ \lambda\sum_{j=1}^J \sum_{v \in\Vleaf} w_v {\Vert
\blbbeta_j^{G_v} \Vert}_2.
\nonumber
\end{eqnarray}
We notice that in the above equation, the first penalty term for $\Vint
$ contains overlapping groups, whereas
the second penalty term for $\Vleaf$ is equivalent to the weighted
lasso penalty
$ \lambda\sum_{j=1}^J \sum_{k=1}^K w_{v(k)} | \beta_j^k|$, where
$w_{v(k)}$ represents
the weight for the leaf node associated with the $k$th response.

Since the penalty term associated with $\Vint$ contains overlapping
groups and therefore
is nonseparable, we rewrite this term by
introducing a vector of auxiliary variables ${\bolds\alpha}_{j}^{G_v}$
for each covariate $j$ and group $G_v$ and by reformulating it with a
dual norm
representation
$\|{\bolds\beta}_{j}^{G_v}\|_2 = \max_{\|{\bolds\alpha
}_{j}^{G_v}\|_2 \leq1} ({\bolds\alpha}_{j}^{G_v})^T{\bolds
\beta}_{j}^{G_v}$ to obtain
%
\begin{eqnarray}
\label{eqpd}
\Omega(\bB)
&\equiv&
\lambda\sum_{j=1}^J \sum_{v\in\Vint} w_v \|{\bolds\beta
}_{j}^{G_v}\|_2
\nonumber\\[-8pt]\\[-8pt]
&= & \lambda\sum_{j=1}^J \sum_{v\in V'} w_v
\max_{\|{\bolds\alpha}_{j}^{G_v}\|_2 \leq1}
({\bolds\alpha}_{j}^{G_v})^T{\bolds\beta}_{j}^{G_v}= \max
_{\bA\in\mathcal{Q}} \langle C\bB^T, \bA
\rangle,\nonumber
\end{eqnarray}
where $\langle\mathbf{U}, \mathbf{V} \rangle\equiv
\Tr(\mathbf{U}^T\mathbf{V})$ denotes a matrix inner product, and
$\bA$ is\break a
$({\sum_{v \in\Vint}}|G_v|) \times J$
matrix given as
%
\[
\bA=\pmatrix{
{\bolds\alpha}_{1}^{G_1}&\cdots&{\bolds\alpha
}_{J}^{G_1}\cr
\vdots&\ddots&\vdots\vspace*{3pt}\cr
{\bolds\alpha}_{1}^{G_{|\Vint|}}&\cdots&{\bolds\alpha
}_{J}^{G_{|\Vint|}}}\vadjust{\goodbreak}
\]
with domain $ \mathcal{Q} \equiv\{\bA| \|{\bolds\alpha
}_{j}^{G_v}\|_2
\leq1 , \forall j \in\{1,\ldots, J\} , v \in\Vint
\}$. In addition,\break $C$~in~(\ref{eqpd}) is a $({\sum_{v \in
\Vint}}|G_v|) \times K$
matrix whose elements are defined as
\[
C_{(v,i)}^k = \cases{
\lambda w_v, &\quad if $k \in G_v$, \cr
0, &\quad otherwise,} 
%
\]
with rows indexed by $(v,i)$ such that $v \in\Vint$ and $i \in G_v$,
and columns indexed by $k \in\{1,\ldots,K \}$. We note\vspace*{1pt}
that the nonseparable terms over ${\bolds\beta }_j^{G_v}$'s in the
tree-lasso penalty are decoupled in the dual-norm representation in~(\ref{eqpd}).

\subsubsection{Smooth approximation to the nonsmooth penalty}

The reformulation in (\ref{eqpd}) is still nonsmooth in $\bB$,
which makes it nontrivial to optimize. To overcome this challenge,
SPG introduces a smooth approximation of (\ref{eqpd}) as follows:
%
\begin{equation}
\label{eqfmut}
f_\mu(\bB)=\max_{\bA\in\mathcal{Q}} \langle C\bB^T, \bA\rangle
- \mu d (\bA),
\end{equation}
where $d(\bA) \equiv\frac{1}{2}\|\bA\|_F^2$ is a smoothing function
with the maximum value $D \equiv\max_{\bA\in\mathcal{Q}} d(\bA) =
\frac{J|\Vint|}{2}$,
and $\mu$ is the parameter that determines the amount of smoothness.
We notice that when $\mu=0$, we recover the original nonsmooth penalty
in $f_0(\bB)$.
It has been shown [\citet{Chen2011}] that~$f_\mu(\bB)$ is convex and smooth
with gradient
\[
\nabla
f_\mu(\bB)=(\bA^{\ast})^TC,
\]
where $\bA^{\ast}$ is the optimal solution to (\ref
{eqfmut}), composed
of $({\bolds\alpha}_j^{G_v})^{\ast} = S(\frac{\lambda w_v
{\bolds\beta}_{j}^{G_v}}{\mu})$,
given the shrinkage operator $S(\cdot)$ defined as
%
\begin{equation}
S(\bu) = \cases{\displaystyle \frac{\bu}{\|\bu\|_2}, &\quad if $\|\bu\|_2
> 1$,\vspace*{2pt}\cr
\bu, &\quad if $\|\bu\|_2 \leq1$.}
\end{equation}
In addition, $\nabla f_\mu(\bB)$ is Lipschitz continuous with the
Lipschitz constant $L_\mu=\|C\|^2/\mu$, where $\|C\| \equiv
\max_{\|\bV\|_F \leq1} \|C\bV^T\|_F$ is a matrix spectral norm. We can
show that $\|C\|= \lambda\max_{k \in\{1, \ldots, K\}}
\sqrt{\sum_{ v\in\Vint\ \mathrm{s.t.}\ k\in G_v}(w_v)^2}$.

\subsubsection{Smoothing proximal gradient (SPG) method}
By substituting the penal\-ty term for $\Vint$ in (\ref
{eqpsplit}) with
$f_\mu(\bB)$ in (\ref{eqfmut}), we obtain an objective
function whose
nonsmooth component contains only the weighted lasso penalty as follows:
%
\begin{equation}\label{eqpsmooth}
\hat{\mathbf{B}}^{T} = \argmin_\bB
\frac{1}{2}{\| \mathbf{Y} - \mathbf{X}\mathbf{B} \|}_F^2
+ f_\mu(\bB)
+ \lambda\sum_{j=1}^J \sum_{k=1}^K w_k |\beta_j^k|.
\end{equation}
The smooth part of the above objective function is
%
\begin{equation}
h(\bB) = {\| \mathbf{Y} - \mathbf{X}\mathbf{B} \|}_F^2 + f_\mu(\bB)\vadjust{\goodbreak}
\end{equation}
and its gradient is given as
%
\begin{equation}\label{eqfgrad}
\nabla h (\bB)= \bX^T(\bX\bB- \bY) + {(\bA^{\ast})}^T C,
\end{equation}
which is Lipschitz-continuous with the Lipschitz constant,
%
\begin{equation}
\label{eqL}
L=\lambda_{\max} (\bX^T\bX)+ L_{\mu} = \lambda_{\max} (\bX^T\bX
) +
\frac{\|C\|^2}{\mu},
\end{equation}
where $\lambda_{\max} (\bX^T\bX)$ is the largest eigenvalue of
$(\bX^T\bX)$.

\begin{algorithm}[t]
\begin{flushleft}
\caption{Smoothing proximal gradient descent (SPG) for tree lasso}
\label{algospg}
\textbf{Input}: $\bX$, $\bY$, $C$, $\bB^0$, Lipschitz constant $L$, desired accuracy $\varepsilon$.

\textbf{Initialization}: set $\mu=\frac{\varepsilon}{2D}$
where $D=\max_{\bA \in \mathcal{Q}} \frac{1}{2}\|\bA\|_F^2=J|V_{\mathrm{int}}|/2$, $\theta_0=1$, $\bW^0=\bB^0$.

\textbf{Iterate} For $t=0,1,2,\ldots,$ until convergence of $\bB^t$:
\begin{enumerate}[\textbf{aaIte}]
\item Compute $\nabla h(\bW^t)$ according to
(\ref{eqfgrad}).
\item Solve the proximal operator associated with the $\ell_1$-norm:
%
\begin{eqnarray}
\label{eqgradupdate}
\bB^{t+1}&=& \argmin_{\bB} Q_L(\bB, \bW^t) \nonumber\\[-8pt]\\[-8pt]
&\equiv& h(\bW^t)+ \langle
\bB-\bW^t, \nabla h(\bW^t) \rangle
+ \lambda \|\bB\|_1
+\frac{L}{2}\|\bB-\bW^t\|_2^2.\nonumber
\end{eqnarray}
\item Set $\theta_{t+1}=\frac{2}{t+3}$.
\item Set $ \bW^{t+1} = \bB^{t+1}+\frac{1-\theta_t}{\theta_t}\theta_{t+1}(\bB^{t+1}-\bB^t)$.
\end{enumerate}
\textbf{Output}: $\hat{\bB} =\bB^{t+1}$.
\end{flushleft}
\end{algorithm}

The key idea behind SPG is that once we introduce the smooth approximation
of (\ref{eqpd}), the only nonsmooth component in
(\ref{eqpsmooth}) is
the weighted lasso penalty and FISTA can be adopted to optimize
(\ref{eqpsmooth}). The
SPG algorithm for the tree lasso is given in Algorithm~\ref{algospg}.
In order to obtain the proximal operator associated with the weighted
lasso penalty,
we rewrite $Q_L(\bB, \bW^t)$ in (\ref{eqgradupdate}) as follows:
\[
Q_L(\bB, \bW^t)= \frac{1}{2}\biggl\|\bB-\biggl(\bW^t-\frac{1}{L}\nabla
h(\bW^t)\biggr)\biggr\|_2^2 + \frac{\lambda}{L} \sum_{j=1}^J
\sum_{k=1}^K w_{v(k)} |\beta_j^k|,
\]
and obtain the closed-form solution for $\bB^{t+1}$ in
(\ref{eqgradupdate})
by soft-thresholding:
\[
\beta_j^k = \operatorname{sign}(v_j^k) \max\biggl(0,
|v_j^k|-\frac{\lambda w_{v(k)}}{L}\biggr),\qquad j=1, \ldots, J \mbox{
and } k=1,\ldots, K,
\]
where $v_j^k$'s are elements of $\bV=(\bW^t-\frac{1}{L}\nabla h(\bW^t))$.
The Lipschitz constant~$L$ given as in (\ref{eqL}) plays the
role of
determining the step size in each gradient descent iteration,
although this value can be expensive to compute for large~$J$.\vadjust{\goodbreak}
As suggested in \citet{Chen2011}, a back-tracking line search
can be used
to determine the step size for large $J$ [\citet{Boyd2004}].

It can be shown that the convergence rate of Algorithm~\ref{algospg}
is $O(\frac{1}{\varepsilon})$ iterations, given the desired accuracy
$\varepsilon$ [\citet{Chen2011}]. If we precompute and store $\bX^T\bX$ and
$\bX^T\bY$, the time complexity per iteration of SPG for the tree lasso
is $O(J^2K+ J\sum_{v\in V}|G_v|)$, compared to $O(J^2
(K+|V_{\mathrm{int}}|)^2 (KN+J(|V_{\mathrm{int}}|+\sum_{v\in
V}|G_v|)))$ for the interior point method for the second-order
cone program. Thus, the time complexity for SPG is quadratic in $J$ and
linear in max($K$, $\sum_{v \in V} |G_v|$), which is significantly more
efficient than cubic in both $J$ and $K$ for the interior point
method.

\section{Experiments}\label{sec4}

We demonstrate the performance of our method on simulated data sets and the
yeast data set of genotypes and gene expressions, and compare the
results with those from the lasso and the $L_1/L_2$-regularized
multi-task regression
that do not assume any structure over responses.
In all of our experiments, we determine the regularization parameter
$\lambda$
by fitting models on a training set for a range of values for $\lambda$,
computing the prediction error of each model on a validation set,
and then selecting the value of a regularization parameter
that gives the lowest prediction error.
We evaluate these methods based on two criteria,
sensitivity/specificity in detecting true relevant covariates and
prediction errors on test data sets.
We note that the $1-$ (specificity) and sensitivity are equivalent to
type I error rate and
$1-$ (type II error rate), respectively.
Test errors are obtained as mean squared differences between the
predicted and observed
response measurements based on test data sets that are independent of
training and validation data sets.

\subsection{Simulation study}\label{sec41}

\begin{figure}
\begin{tabular}{@{}ccc@{}}

\includegraphics{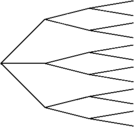}
 & \includegraphics{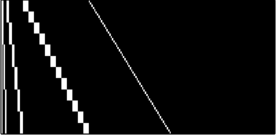} & \includegraphics{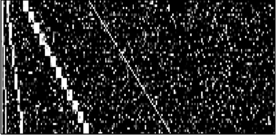}\\
(a) & (b) & (c)\\[4pt]
&
\includegraphics{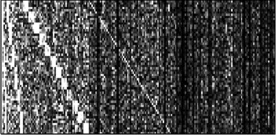}
 & \includegraphics{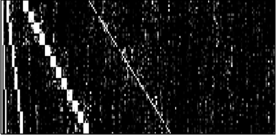}\\
& (d) & (e)
\end{tabular}
\caption{An example of regression coefficients estimated from a
simulated data set. \textup{(a)}:~Hierarchical clustering tree of four levels over
responses. Only the top three levels are shown to avoid clutter.
\textup{(b)}:~True regression coefficients. Estimated parameters are shown for \textup{(c)}:
lasso, \textup{(d)}:~$L_1/L_2$-regularized multli-task regression, and \textup{(e)}: tree
lasso. The rows represent responses and the columns covariates.}
\label{figsimb}
\end{figure}

We simulate data using the following scenario analogous to eQTL
mapping. We simulate $(\mathbf{X},\mathbf{Y})$ with $K = 60$, $J =
200$, and $N = 150$ as follows. We first generate
the genotypes $\mathbf{X}$ by sampling each element in $\mathbf{X}$
from a uniform distribution
over $\{0, 1, 2\}$ that corresponds to the number of mutated alleles
at each SNP locus. Then, we set
the values of $\mathbf{B}$ by first selecting nonzero entries
and filling these entries with predefined values.
We assume a hierarchical structure with four levels over the responses,
and select the nonzero elements of $\mathbf{B}$ so that
the groups of responses described by the tree share common relevant covariates.
The hierarchical clustering tree as used in our simulation is shown in
Figure~\ref{figsimb}(a)
only for the top three levels to avoid a clutter, and the true
nonzero elements in the regression coefficient matrix are shown as
white pixels
in Figure~\ref{figsimb}(b) with responses (gene expressions) as rows and
covariates (SNPs) as columns.
In all of our simulation study,
we divide the full data set of $N=150$ into training and validation
sets of sizes 100 and 50, respectively.

%

%

To illustrate the behavior of different methods,
we fit the lasso, the $L_1/L_2$-regularized multi-task regression,\vadjust{\goodbreak} and
our method to
a single data set simulated with the nonzero elements of $\mathbf{B}$
set to 0.4, and show the results in Figure~\ref{figsimb}(c)--(e), respectively.
Since the lasso does not have any mechanism to borrow statistical
strength across different responses,
false positives for nonzero regression coefficients are distributed
randomly across the matrix~$\hat{\mathbf{B}}^{\mathrm{lasso}}$ in
Figure~\ref{figsimb}(c).
On the other hand, the $L_1/L_2$-regularization method
blindly combines information across all responses regardless of the correlation
structure. As a result, once a covariate is selected as relevant for a~response,
it gets selected for all of the other responses, and we observe vertical
stripes of nonzero values in Figure~\ref{figsimb}(d). When
the hierarchical clustering structure in Figure~\ref{figsimb}(a) is available
as prior knowledge, it is visually clear from Figure~\ref{figsimb}(e)
that our method is able to suppress false positives,
and to recover the true relevant covariates for correlated responses
significantly
better than other methods.

%
\begin{figure}
\begin{tabular}{@{}c@{\qquad}c@{}}

\includegraphics{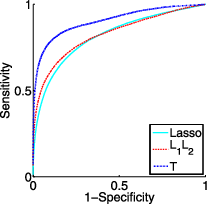}
 & \includegraphics{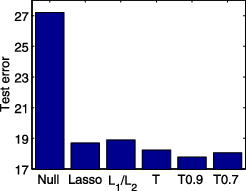}\\
(a) & (b)
\end{tabular}
\caption{Comparison of various sparse regression methods on simulated
data sets. \textup{(a)}:~ROC curves for the recovery of true relevant covariates.
\textup{(b)}: Prediction errors. In simulation, $\beta^j_{k}=0.2$ is used for the
nonzero elements of the true regression coefficient matrix. Results are
averaged over 50 simulated data sets.}
\label{figsimroc}
\end{figure}

In order to systematically evaluate the performance of the different
methods, we generate
50 simulated data sets, and show in Figure~\ref{figsimroc}(a)
receiver operating characteristic (ROC) curves for the
recovery of the true nonzero elements in the regression coefficient
matrix averaged over these 50 data sets.
Figure~\ref{figsimroc}(a) represents results from data sets with
true nonzero elements in $\mathbf{B}$ set to 0.2. Additional results
for true nonzero elements in $\mathbf{B}$ set to 0.4 and 0.6 are
available in Online Appendix Figures 1A and 1B
[\citet{Kim12}].
Our method clearly outperforms the lasso and the $L_1/L_2$-regularized
multi-task regression.
Especially when the signal-to-noise ratio is low in Figure~\ref{figsimroc}(a),
the advantage of incorporating the prior knowledge of the tree as a
correlation structure
over responses is significant.

We compare the performance of the different methods in terms of prediction
errors, using an additional 50 samples as test data. The prediction
errors averaged over 50
simulated data sets are shown in Figure~\ref{figsimroc}(b) for data sets
generated from 0.2 for true nonzero elements of regression coefficients.
Additional results for data sets generated from 0.4 and 0.6 for true
nonzero elements of regression coefficients
are shown in Online Appendix Figures~2A and~2B, respectively.
In addition to the results from sparse regression methods,
we include the prediction errors from the null model that has only an
intercept term.
We find that our method shown as ``T'' in Figure~\ref{figsimroc}(b) has
lower prediction errors than
all of the other methods.
In the tree lasso, in addition to directly using the true tree
structure in
Figure~\ref{figsimb}(a),
we also consider the scenario in which the true tree structure 
is not known a priori. In this case, we learn a tree
by running a hierarchical agglomerative clustering algorithm
on the $K\times K$ correlation matrix
of the response measurements, and use this tree along with the weights
$h_v$'s associated with
each internal node in our method.
Since the tree obtained in this manner represents a~noisy realization
of the true underlying tree structure, we
discard the nodes for weak correlation near the root of the tree by
thresholding the normalized~$h_v$'s at $\rho=0.9$ and 0.7, and show
the prediction errors
obtained from these thresholded trees
as ``T0.9'' and ``T0.7'' in Figure~\ref{figsimroc}(b).
Even when the true tree structure is not available, our method is able
to benefit from taking into account
the correlation structure among responses, and gives lower prediction errors.
We performed the same experiment while varying the threshold~$\rho$ in
the range of [0.6, 1.0],
and obtained similar prediction errors across different values of $\rho
$ (results not shown).
This shows that the meaningful clustering information that the tree
lasso takes advantage of lies mostly
in the tight clusters at the lower levels of a tree rather than the clusters
of loosely related variables near the root of the tree.

\subsection{Analysis of yeast data}\label{sec42}

We analyze the yeast eQTL data set of the genotype and gene-expression
data for 114 yeast strains
[\citet{brem2008}] using various sparse regression methods. We focus
on the chromosome 3 with 21 SNPs and expression levels of 3,684 genes,
after removing those genes whose expression levels are missing in more
than 5\% of the samples.
Although it is widely known that genes are organized into functional
modules within
which gene-expression levels are often correlated,
the hierarchical module structure over correlated genes is not
directly available as prior knowledge, and we learn the tree
by running the hierarchical agglomerative clustering
algorithm on gene-expression data. We
use only the internal nodes with heights $h_v<0.7$ or $0.9$ in our method.
The goal of the analysis is to search for SNPs (covariates) whose variation
induces a significant variation
in the gene-expression levels (responses) over different strains. By applying
our method that incorporates information on gene modules at multiple granularity
in the hierarchical clustering tree, we expect to be able to identify
SNPs that influence the activity of a group of genes that are
co-expressed or co-regulated.

%
\begin{figure}[b]
\begin{tabular}{@{}c@{\hspace*{4pt}}c@{\hspace*{4pt}}
c@{\hspace*{4pt}}c@{\hspace*{4pt}}c@{}}

\includegraphics{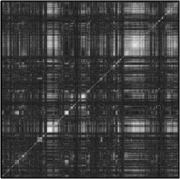}&
\includegraphics{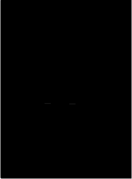} &
\includegraphics{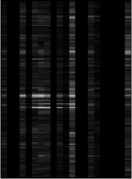} &
\includegraphics{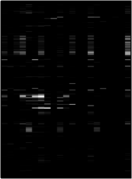}
&
\includegraphics{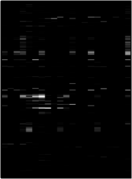}
\\
(a) & (b) & (c) & (d) & (e)
\end{tabular}
\caption{Results for the yeast eQTL data set. \textup{(a)}: Correlation matrix of
the gene expression data, where rows and columns are reordered after
applying hierarchical agglomerative clustering. Estimated regression
coefficients are shown for \textup{(b)}: lasso, \textup{(c)}: $L_1/L_2$-regularized
multi-task regression, \textup{(d)}: tree lasso with $\rho=0.9$, and \textup{(e)}: with
$\rho=0.7$. In panels \textup{(b)}--\textup{(e)}, the rows represent genes (responses) and the
columns SNPs (covariates).}
\label{figyb}
\end{figure}

\begin{figure}

\includegraphics{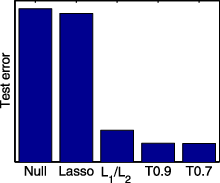}

\caption{Prediction errors for the yeast eQTL data set.}
\label{figytserr}\vspace*{-3pt}
\end{figure}

In Figure~\ref{figyb}(a), we show the $K \times K$ correlation matrix
of the gene expressions after reordering the rows and columns according
to the results of the hierarchical agglomerative clustering algorithm.
The estimated $\mathbf{B}$
is shown for the lasso, the $L_1/L_2$-regularized multi-task
regression, and our method
with $\rho=0.9$ and 0.7 in Figure~\ref{figyb}(b)--(e), respectively,
where the rows represent genes and the columns SNPs.
The regularization parameter is chosen based on prediction errors on
a validation set of size 10.
The lasso estimates in Figure~\ref{figyb}(b) are extremely sparse
and do not reveal any interesting structure in SNP-gene relationships.
We believe that the association signals are very weak as is typically
the case in the eQTL study, and that the lasso is
unable to detect such weak signals without combining statistical
strength across multiple\vadjust{\goodbreak}
genes with correlated expressions.
The estimates from the $L_1/L_2$-regularized multi-task regression
are not sparse across gene expressions, and tend to form vertical
stripes of
nonzero regression coefficients as can be seen in Figure~\ref{figyb}(c).
On the other hand, our method in Figure~\ref{figyb}(d)--(e) reveals clear groupings
in the patterns of associations between gene expressions and SNPs. In addition,
as shown in Figure~\ref{figytserr}, our method
performs significantly better in terms of prediction errors on the test set
obtained from the 10-fold cross-validation.

%

Given the estimates of $\mathbf{B}$ in Figure~\ref{figyb},
we look for an enrichment of gene ontology (GO) categories among the
genes with
nonzero estimated regression coefficients for each SNP. A group of genes
that form a module often participate in the same
pathway, leading to an enrichment of a GO category among the members
of the module. Since we are interested in identifying SNPs influencing
gene modules,
and our method
encourages this joint association through the hierarchical clustering tree,
we hypothesize that our method would reveal more significant GO enrichments
in the estimated nonzero elements in $\mathbf{B}$.
Given the tree-lasso estimate,
we search for GO enrichment in the set of genes that have nonzero
regression coefficients
for each SNP.
On the other hand, the estimates of the $L_1/L_2$-regularized method
are not sparse across genes. Thus,
we threshold the absolute values of the estimated~$\mathbf{B}$ at 0.005,
0.01, 0.03, and 0.05, and perform GO enrichment analysis for only those
genes with $\beta^k_j$ above the threshold.

\begin{figure}
\begin{tabular}{@{}ccc@{}}

\includegraphics{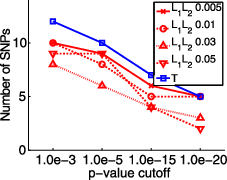}
 & \includegraphics{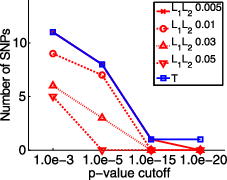} & \includegraphics{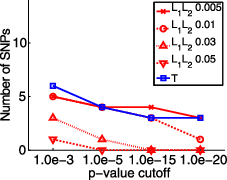}\\
(a) & (b) & (c)
\end{tabular}
\caption{Enrichment of GO categories for genes whose expression-levels
are influenced by the same SNP based on the regression coefficients
estimated from the yeast eQTL data set. The number of SNPs with
significant enrichment is shown for GO categories within \textup{(a)}:~biological
process, \textup{(b)}: molecular function, and \textup{(c)}: cellular component.}
\label{figygo}
\end{figure}

In Figure~\ref{figygo}, we show the number of SNPs with significant enrichments
at different $p$-value cutoffs
for subcategories within each of the three broad GO categories,
including biological processes, molecular functions, and cellular components.
For example, within biological processes, SNPs were found to be
enriched for GO terms such as
mitocondrial translation, amino acid biosynthetic process, and organic
acid metabolism.
Regardless of the thresholds for selecting significant
associations in the estimates from the $L_1/L_2$-regularized multi-task
regression,
our method generally finds more significant enrichment.
Although due to the lack of ground-truth information,
the results in Figure~\ref{figygo} do not directly demonstrate
that our method led to more significant findings than other methods,
they provide evidence that our method was successful in finding
SNPs with pleiotropic\vadjust{\goodbreak} effects that influence gene modules rather than
focusing on identifying SNPs that affect individual genes as in the lasso.

%

%
\begin{table}[!t]
\tabcolsep=0pt
\caption{Enriched GO categories for genes whose expression levels are influenced by
the~same~SNP in the yeast eQTL data set. The results in columns 1--4 are
based on~the~tree-lasso estimate
of regression coefficients. The last column shows the~enriched~GO
categories reported in Zhu et al. (\protect\citeyear{brem2008})
(BP:~biological processes, MF:~molecular functions, CC: cellular components)}
\label{tblygo}
{\fontsize{8.68pt}{11pt}\selectfont{
\begin{tabular*}{\tablewidth}{@{\extracolsep{\fill}}@{}lclll@{}}
\hline
&&&& \multicolumn{1}{c@{}}{\textbf{Previously}}\\
\textbf{SNP} &  & &  & \multicolumn{1}{c@{}}{\textbf{reported}}\\
\textbf{loc.} & \textbf{Module} & \multicolumn{1}{c}{\textbf{GO category}} & &  \multicolumn{1}{c@{}}{\textbf{enrichment [Zhu}}\\
\textbf{in Chr3} & \textbf{size} & \multicolumn{1}{c}{\textbf{(overlap/\#genes)}} & $\bolds{p}$\textbf{-value}
& \multicolumn{1}{c@{}}{\textbf{et al. (\citeyear{brem2008})]}}\\
\hline
\hphantom{0}64,300 & 203 & BP: Amino acid biosynthetic process $(36/92)$ & $3.8\times
10^{-20}$ & \\
[4pt]
\hphantom{0}75,000 & 167 & BP: Amino acid biosynthetic process $(46/92)$ & $8.7\times
10^{-37}$ & BP: Organic \\
& & BP: Organic acid metabolism $(62/244)$ & $2.6\times10^{-30}$ & acid
metabolism \\
& & MF: Transferase activity $(47/476)$ & $7.0\times10^{-6}$ &
($1.6\times10^{-42}$) \\
[4pt]
\hphantom{0}76,100 & 186 & MF: Catalytic activity $(106/1379)$ & $3.3\times10^{-6}$
& \\
[4pt]
\hphantom{0}79,000 & 167 & BP: Amino acid biosynthetic process $(52/92)$ & $6.1\times
10^{-46}$ & \\
& & MF: Catalytic activity $(99/1379)$ & $5.4\times10^{-7}$ & \\
[4pt]
\hphantom{0}86,000 & 103 & BP: Amino acid biosynthetic process $(29/92)$ & $6.3\times
10^{-22}$ & \\
& & MF: Oxidoreductase activity $(20/197)$ & $2.3\times10^{-5}$ & \\
[4pt]
100,200 & \hphantom{0}68 & BP: Amino acid biosynthetic process $(19/92)$ & $1.4\times
10^{-13}$ & \\
[4pt]
105,000 & 168 & BP: Amino acid biosynthetic process $(45/92)$ &
$3.2\times10^{-35}$ & \\
& & MF: Transferase activity $(47/476)$ & $1.0\times10^{-5}$ & \\
[4pt]
175,800 & \hphantom{0}89 & BP: Amino acid biosynthetic process $(34/92)$ & $1.7\times
10^{-31}$ & \\
& & MF: Catalytic activity $(59/1379)$ & $2.1\times10^{-6}$ & \\
[4pt]
210,700 & \hphantom{0}23 & BP: Branched chain family & $3.4\times10^{-9}$ & BP:
Response to \\
& & amino acid biosynthetic process $(6/12)$ & & chemical
stimulus \\
& & BP: Response to pheromone $(8/69)$ & $4.1\times10^{-8}$ &
($7.6\times10^{-7}$) \\
[4pt]
228,100 & 195 & BP: Mitochondrial translation $(32/77)$ & $2.9\times
10^{-19}$ & \\
& & CC: Mitochondrial part $(77/345)$ & $9.3\times10^{-30}$ & \\
& & MF: Hydrogen ion transporting ATP synthase & $3.3\times10^{-10}$ &
\\
& & activity, rotational mechanism $(9/9)$ & & \\
[4pt]
240,300 & 258 & CC: Cytosolic ribosome $(110/140)$ & $9.6\times
10^{-107}$ & \\
& & MF: Structural constituent of ribosome & $8.1\times10^{-75}$ & \\
& & $(104/189)$ & & \\
[4pt]
240,300 & \hphantom{0}40 & BP: Generation of precursor & $6.1\times10^{-13}$ & \\
& & metabolites and energy $(17/132)$ & & \\
& & CC: Mitochondrial inner membrane $(13/126)$ & $1.7\times10^{-8}$ & \\
& & MF: Transmembrane transporter activity & $2.8\times10^{-7}$ & \\
& & $(14/195)$ & & \\
[4pt]
301,400 & 274 & MF: snoRNA binding $(13/16)$ & $1.0\times10^{-10}$ \\
\hline
\end{tabular*}}}
\end{table}

Table~\ref{tblygo} lists the enriched GO categories ($p$-$\mbox{value} <
1.0\times10^{-5}$)
for SNPs and the groups of genes whose expression levels are affected
by the given SNP
based on the tree-lasso estimate of association strengths.
For comparison, in the last column of Table~\ref{tblygo},
we include the enriched GO categories for roughly similar genomic locations
that have been previously reported in \citet{brem2008} using the conventional
single-SNP/single-gene statistical test for association.
While the tree-lasso results mostly recover the previously-reported GO
enrichments, we
find many additional enrichments that are statistically significant.
This observation again provides us with indirect evidence that the tree
lasso can extract
fine-grained information on gene modules perturbed by genetic polymorphisms.

\section{Discussion}\label{sec5}

In this article we proposed a novel regularized regression approach,
called the tree lasso, that identifies covariates relevant to multiple
related responses jointly by leveraging the correlation structure
in responses represented as a hierarchical clustering tree.
We discussed how this approach can be used in eQTL analysis to learn
SNPs with pleiotropoic effects that influence the activities of
multiple co-expressed genes.
For optimization, we adopted the smoothing proximal gradient approach
that was originally developed for a general class of
structured-sparsity-inducing penalties,
as the tree-lasso penalty can be viewed as a special case.
Our results on both the simulated and yeast data sets showed a clear advantage
of the tree lasso in increasing the power of detecting weak signals and
reducing false positives.

\begin{supplement}
\stitle{The balanced weighting scheme of tree lasso and additional
experimental results}
\slink[doi]{10.1214/12-AOAS549SUPP} 
\slink[url]{http://lib.stat.cmu.edu/aoas/549/supplement.pdf}
\sdatatype{.pdf}
\sdescription{We prove that the weighting scheme of the tree-lasso
penalty achieves a balanced penalization
of all regression coefficients. We also provide additional experimental
results on
the comparison of the tree lasso with other sparse regression methods
using simulated data sets.}
\end{supplement}

%

\printaddresses

\end{document}